\documentclass[10pt,twocolumn,letterpaper]{article}

\usepackage{titling}
\usepackage{iccv}
\usepackage{times}
\usepackage{epsfig}
\usepackage{graphicx}
\usepackage{amsmath}
\usepackage{amssymb}
\usepackage{multirow}
\usepackage{booktabs}

\usepackage[breaklinks=true,bookmarks=false]{hyperref}

\iccvfinalcopy

\begin{document}

\title{\emph{DeViL:} Decoding Vision features into Language}

\date{\vspace{-4mm}}
\author{
 Meghal Dani$^{*1}$ \quad 
 Isabel Rio-Torto$^{*3,4}$  \quad
 Stephan Alaniz$^{1}$  \quad  
 Zeynep Akata$^{1,2}$
\\\\
$^{1}$University of Tübingen \quad
$^{2}$ MPI for Intelligent Systems \quad
$^{3}$ INESC TEC \\
$^{4}$Faculdade de Ciências da Universidade do Porto \\
$^{*}$ Equal Contribution
}

\maketitle
\ificcvfinal\thispagestyle{empty}\fi

\begin{abstract}
Post-hoc explanation methods have often been criticised for abstracting away the decision-making process of deep neural networks. In this work, we would like to provide natural language descriptions for what different layers of a vision backbone have learned. Our \emph{DeViL} method decodes vision features into language, not only highlighting the attribution locations but also generating textual descriptions of visual features at different layers of the network. We train a transformer network to translate individual image features of any vision layer into a prompt that a separate off-the-shelf language model decodes into natural language. By employing dropout both per-layer and per-spatial-location, our model can generalize training on image-text pairs to generate localized explanations. As it uses a pre-trained language model, our approach is fast to train, can be applied to any vision backbone, and produces textual descriptions at different layers of the vision network. Moreover, \emph{DeViL} can create open-vocabulary attribution maps corresponding to words or phrases even outside the training scope of the vision model. We demonstrate that \emph{DeViL} generates textual descriptions relevant to the image content on CC3M surpassing previous lightweight captioning models and attribution maps uncovering the learned concepts of the vision backbone. Finally, we show \emph{DeViL} also outperforms the current state-of-the-art on the neuron-wise descriptions of the MILANNOTATIONS dataset.
Code available at {\small{\url{https://github.com/ExplainableML/DeViL}}}.
\end{abstract}

\section{Introduction}

\begin{figure}[t]
  \includegraphics[width=\linewidth]{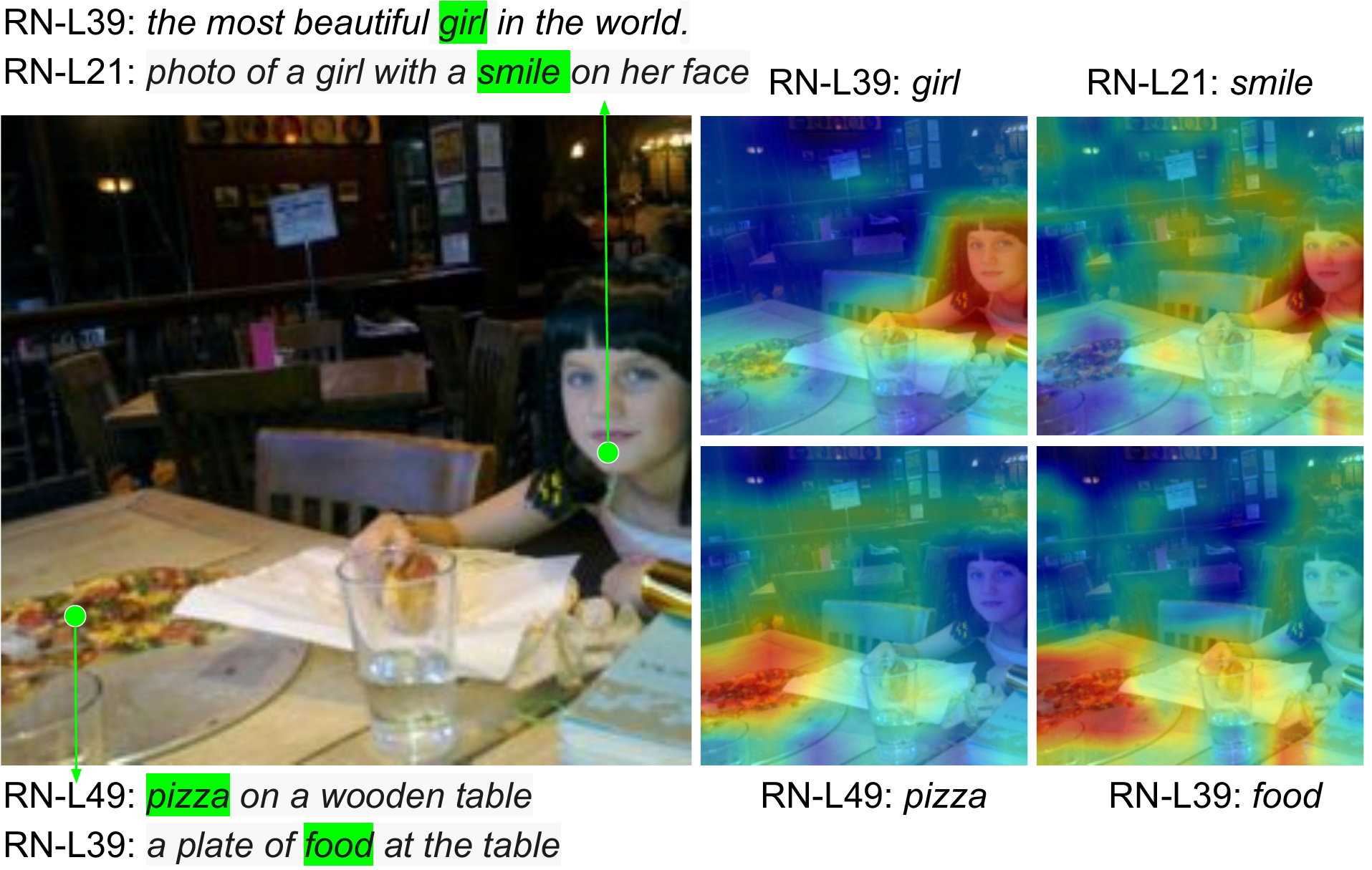}
  \caption{Our \emph{DeViL} model generates textual descriptions for feature maps of any arbitrary vision backbone. We show the attribution maps for \textit{girl}, \textit{smile}, \textit{pizza} and \textit{food} concepts generated by the layer 49 (RN-L49), layer 39 (RN-L39) and layer 21 (RN-L21) of a ResNet-50 backbone respectively. We select these words from the sentences that are generated for these layer outputs (highlighted in green).}
  \label{fig:teaser}
\end{figure}

Despite the significant progress and success of deep vision models, widespread adoption in safety-critical fields, such as application in medicine, is inhibited by the lack of understanding and interpretation of how deep models arrive at their predictions. Practitioners would like to inspect how a vision model processes its input without making changes to the training procedure such that they can use the latest state-of-the-art models for their domain. For such purposes, posthoc explanation methods allow the interpretation of arbitrary vision models a posteriori without making modifications to the architecture or loss function.

One popular strategy to improve the interpretability of vision models is to generate visual attribution maps. Activation-based methods~\cite{zhou2016learning,selvaraju2017grad,smilkov2017smoothgrad,zeiler2014visualizing, chattopadhay2018grad} take into account the features a vision model produced at a given layer for visualizing the spatial attribution with respect to the network output, e.g., the predicted class for recognition models. However, visualizing saliency maps is just one possible dimension to explaining a vision model and they come with the limitation of not always being reliable~\cite{AdebayoGMGHK18}. Natural language explanations (NLEs) have been proposed in the context of vision-language models to extend the language output for a given task with a fitting explanation~\cite{SammaniMD22}, such as for a VQA task. A less explored direction is directly explaining internal features or neurons of neural networks through natural language~\cite{hernandez2022milan}. Such explanations would allow a wider reach of application, especially where users are not expected to have expert knowledge and can more easily interpret textual explanations than saliency maps.

In this work, we propose \emph{DeViL} (Decoding Vision features into Language) that can explain
 features of a vision model through natural language. By employing a general purpose language model, we inherit rich generalization properties, allowing us to additionally perform open-vocabulary saliency analysis. Thus, \emph{DeViL} combines providing textual explanations of these visual features with highlighting visual attributions spatially to extract diverse explanations that complement each other. Thus, enabling non-experts to comprehend the network's decision-making process, diagnose potential issues in the model's performance, and enhance user trust and accountability.

Figure~\ref{fig:teaser} illustrates the explanations obtained by our \emph{DeViL} method when applied to an ImageNet-trained ResNet50 model. If we want to understand what the network encodes at a specific location of the image, \emph{DeViL} translates the nearest image feature of a layer into language, e.g. at the bottom left the network encodes ``pizza on a wooden table'' at the last convolutional layer, and ``a plate of food at the table'' at a lower layer. Subsequently, we can inspect the network in a broader sense, by querying the keywords ``pizza'' and ``food'' at the respective layers for a more complete explanation. Evidently, layer 49 understands fine-grained concepts while an earlier layer produces a general word ``food''. Similar observations have been previously discovered~\cite{szegedy2015going, zeiler2014visualizing, agrawal2015learning, he2016resnet} and \emph{DeViL} allows us to expose these findings through explorative analysis to users.

With our model \emph{DeViL}, our key contributions are i) generating spatial and layer-wise natural language explanations for any off-the-shelf vision model, ii) open-vocabulary saliency attributions coherent with textual explanations, and iii) showing how dropout can be used to generalize from image-text pairs on a global level to fine-grained network inspections. To the best of our knowledge, this is the first work to combine these capabilities into a single model while requiring short training times and using abundantly available captioning data instead of explanation specific datasets.

\section{Related Work}\label{related_work}

\textbf{Inherently interpretable models.} Apart from post-hoc explanation methods, inherently interpretable models or intrinsic explanation methods try to make modifications to the network architecture or loss function to make the deep model more interpretable. For instance, induced alignment pressure on the weights during optimisation either in a linear~\cite{chen2019looks, brendel2019approximating, bohle2021convolutional, bohle2022b} or non-linear manner~\cite{liu2018decoupled, zoumpourlis2017non} has been shown to produce more interpretable models. While these methods produce more interpretable models they require adaptation and re-training of existing models and cannot always recover the original task performance. With the increasing size of models, this is not always a scalable solution. Post-hoc methods trade-off the guaranteed faithfulness of the explanation with broader applicability.

\textbf{Post-hoc saliency.} Saliency attribution is most commonly produced through perturbations~\cite{LundbergL17,PetsiukDS18}, gradients~\cite{SundararajanTY17}, activations~\cite{KimWGCWVS18,zeiler2014visualizing,zhou2016learning} or the combination of those~\cite{selvaraju2017grad,chattopadhay2018grad}.
Class Activation Mapping (CAM)~\cite{zhou2016learning} and its successors, such as Grad-CAM~\cite{selvaraju2017grad} and Grad-Cam++~\cite{chattopadhay2018grad} are one of the most popular attribution methods that make use of network activations, similar to our approach, but additionally use the gradients of the target class.
We set ourselves apart from these methods by also providing directly related natural language explanations for the saliency maps.

\textbf{Natural Language Explanations.} NLEs are another alternative for explaining a network in a human understandable way. One of the first works on NLEs~\cite{hendricks2016gve} used a Long short-term memory (LSTM) network with a reinforcement learning-based loss to guide generated image descriptions into more class discriminative explanations. More recently, the availability of large language models (LLMs) has also seen them applied to NLE generation. These approaches are usually divided into predict-explain and explain-predict paradigms: in the first, a language model is finetuned with features and predictions from a pretrained vision model~\cite{kayser2021vil, marasovic2020rvt}, while in the latter a single model is jointly trained to explain and predict~\cite{majumder2022rexc, SammaniMD22, pluster2022ofax}. 

While the aforementioned approaches only provide a single NLE for the whole image, Bau~\etal~\cite{bau2017network} attempts to label individual highest activated neurons in a network by correlating them with pixel-wise annotations. Building upon that, Hernandez~\etal~\cite{hernandez2022milan} collect the MILANNOTATIONS datasets where humans labeled sets of images that showed high activations for individual neurons of different network architectures. By training a vision-language model on this data, individual neurons of different layers can be described in natural language. 

In contrast, \emph{DeViL} can generate both types of NLEs (global and local), without requiring a dataset of human annotated descriptions per neuron or layer. Nonetheless, we show that when our model is trained on MILANNOTATIONS, we obtain a higher correspondence with human descriptions.

\begin{figure*}[ht]
  \includegraphics[width=\linewidth]{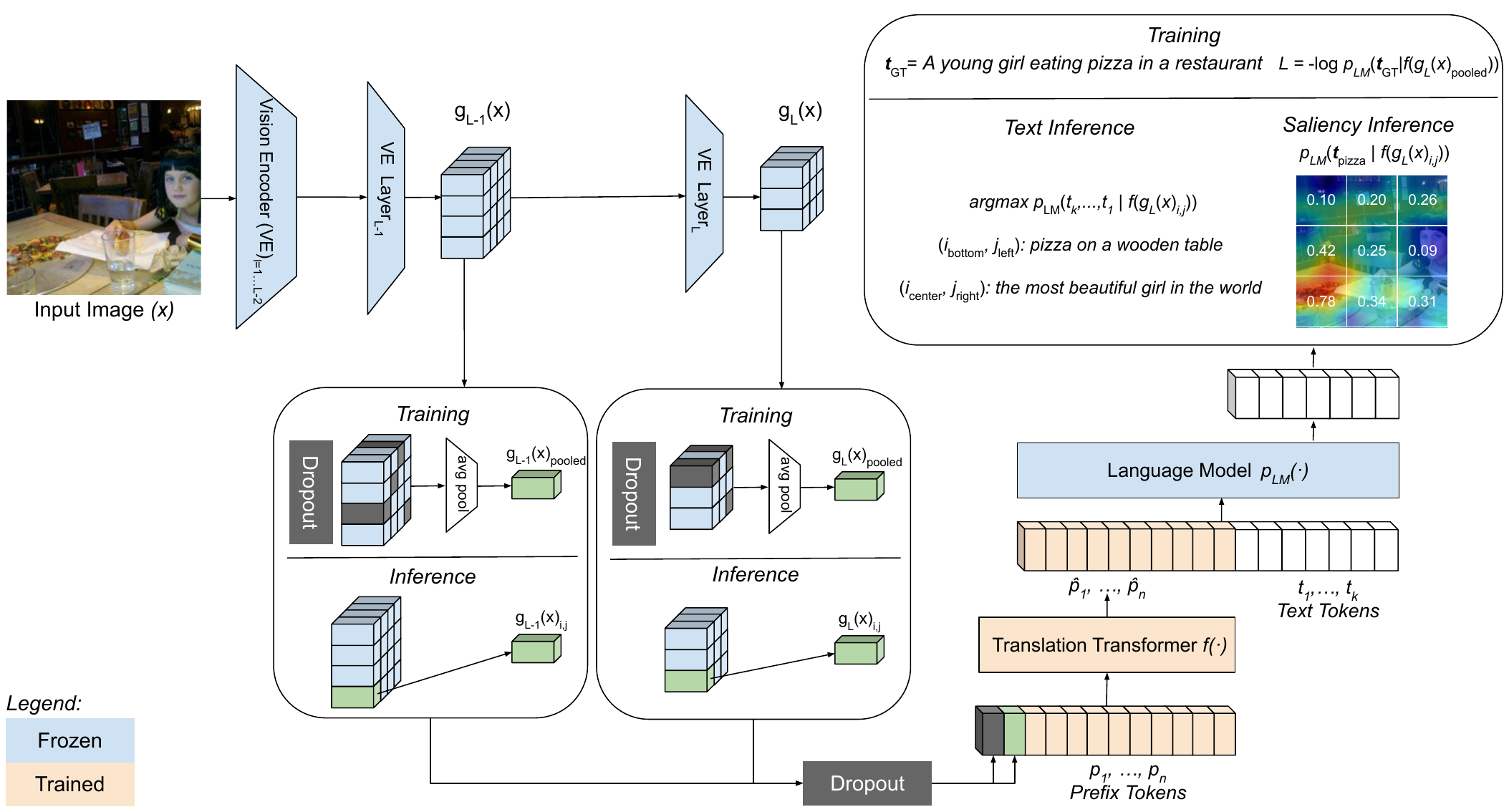}
  \caption{Architecture diagram of our proposed model \emph{DeViL}: Given an input image $x$ and a pretrained vision encoder(VE) with multiple layers($L$), we concatenate the average pooling embedding($g_L(x)_{pooled}$) to learnable prefix tokens  (size=10; as considered by CLipCap~\cite{mokady2021clipcap}). A translation transformer $f$ then projects prefix tokens $\hat{p}_1,\dots,\hat{p}_n$ used to condition language model $p_{\text{LM}}$ to get the captions for the entire layer. During Inference, we can select individual feature $g_L(x)_{i,j}$ for any arbitrary location $i,j$ of the layer and generate its respective NLE, with saliency for any open-vocabulary word. The blocks in blue represent frozen weights while others are trained}
  \label{fig:DeVil}
\end{figure*}

\textbf{Captioning Models.} Image captioning models~\cite{mokady2021clipcap, radford2021learning, tewel2022zerocap,zhou2020unifiedvlp,hu2022lemon}, on the other hand, use LLMs to generate free-form captions for the input image. 
State-of-the-art in captioning is obtained by large-scale pre-training and subsequently finetuning on target datasets~\cite{zhou2020unifiedvlp,hu2022lemon}. On the other hand, lightweight models such as CLIPCap~\cite{mokady2021clipcap} perform competitively by relying on pre-trained vision and language models requiring significantly fewer GPU hours to train.
\emph{DeViL} is similar to CLIPCap in that both models make use of a translation network between image features and the language model prompt. While CLIPCap is capable of generating high-quality captions for images, it cannot be directly used to interpret a model's internal representations. In contrast, by extracting features from multiple layers of the vision model and incorporating dropout, \emph{DeViL} generalizes to more fine-grained descriptions and improves all captioning metrics on CC3M~\cite{sharma2018cc3m}.

\section{\emph{DeViL} Model}\label{method}
Given a pre-trained vision model $g$, our goal is to decode the features of several of its $L$ layers into natural language describing what these features encode. These descriptions are specific to the network's activations corresponding to an input image. Let $g_l(x)$ denote the feature map of the $l$th-layer of $g$ for input image $x$ with spatial size $H_l \times W_l$. We propose \emph{DeViL}, a network that is trained to decode an individual feature vector $g_l(x)_{i,j}$ of an arbitrary spatial location $i,j$ into a natural language description. Since \emph{DeViL} is trained on top of features from a frozen vision model $g$, we view it as a post-hoc explanation method, that is able to reason about which concepts have been learned by the vision model retrospectively.  Figure~\ref{fig:DeVil} depicts an overview of the network architecture of \emph{DeViL}, which we describe in more detail in the following sub-sections.

\subsection{Translating vision features into language}
Our \emph{DeViL} method is designed to be a lightweight model such that it can be trained and applied to existing vision backbones without requiring a long training time. Instead of training a language model, we make use of a pre-trained language model with frozen weights to generate the natural language descriptions for vision features. Specifically, the language model $p_{\text{LM}}(t_k|t_1,\dots,t_{k-1})$ is able to predict the next token $t$ from its tokenizer's vocabulary given a sequence of previous tokens as context. As shown in recent works~\cite{zhou2020unifiedvlp,mokady2021clipcap}, the text generation of language models can be guided by conditioning the model on learned prefix tokens instead of finetuning its weights.

We follow this approach and train only a translation network $f$ that takes image features $g_l(x)_{i,j}$ and produces $n$ prefix tokens $\boldsymbol{\hat{p}} = \{\hat{p}_i\}_{i=1}^n$ that are used to condition the language model $p_{\text{LM}}(t_k|\hat{p}_1,\dots,\hat{p}_n,t_1,\dots,t_{k-1})$ for generating a description. The translation network uses a standard Transformer architecture and takes as input the concatenated sequence of image features $g_l(x)_{i,j}$ and the set of trained parameters for the initial prefix tokens $\boldsymbol{p}$. From the output, we only keep the transformed prefix tokens $\boldsymbol{\hat{p}}$ to pass on to the language model as prompt in a causal language generation setting.

\subsection{Learning to decode individual vision features}
Obtaining fine-grained data on textual descriptions for neurons or activations of the neural networks as done in~\cite{hernandez2022milan} is costly at scale. Hence, we resort to more commonly available text-image pairs in the form of a captioning dataset to achieve the same goal. Since these image-text pairs relate the global image content to a single sentence, we describe in the following how we can train on these global relations, but evaluate our model on more fine-grained feature locations across the layers of the vision model.

Given a vision model $g$, we first specify the layers we would like to decode. For every forward pass of $g$, we extract the features $g_l(x)$ for every layer $l$ in our set of explained layers. At training time, we perform 2D global average pooling over the spatial locations at each feature layer to obtain a single feature vector per layer $g_l(x)_{\text{pooled}}$. After applying a linear projection onto the same dimensionality and adding a positional embedding, the feature vectors are passed to the translation network $f$ as a sequence. Depending on how many layers are being explained, the sequence length for $f$ varies, i.e., it increases by one for each layer that is explained. At inference time, instead of pooling the visual features, we can select a specific location we would like to explain and only pass that particular feature vector $g_l(x)_{i,j}$ to $f$.

To train our translation network, we optimize the standard causal language modelling task of predicting the next token given previous tokens. In our context, the language model is conditioned on the image features through the translation network. Thus, our loss is
\begin{align}
    \mathcal{L} &= \log p_{\text{LM}}(\boldsymbol{t}|f(g(x)))\\
                &= \sum_k \log p_{\text{LM}}(t_k|\hat{p}_1,\dots,\hat{p}_n,t_1,\dots,t_{k-1})
\end{align}
with trainable parameters only in the translation network $f$.
Once trained, \emph{DeViL} can decode vision features into natural language by conditioning the language model on a vision feature vector and generating a sentence. In practice, we greedily choose the most probable next word with $\arg \max$.

\subsection{Generalization through dropout}
The global average pooling at test time is required because we do not have more fine-grained data to train on. However, it creates a discrepancy between training and inference time. To overcome this issue, we introduce two dropout operations. Firstly, we randomly dropout spatial locations ${i,j}$ of $g_l(x)$ before applying average pooling to obtain $g_l(x)_{\text{pooled}}$. As a result, our translation network observes a larger space of image features that better cover the full distribution of a vision layer's features.
Secondly, we randomly subselect the layers from which the translation network $f$ obtains the pooled features. This way, the translation network sometimes receives features only from individual layers during training time. This is crucial because, at inference time, we would like to decode a specific feature of an individual vision layer so this dropout ensures this input configuration is seen during training. Moreover, it allows to train a single network per vision model instead of one per layer.

In contrast to dropout as proposed by~\cite{SrivastavaHKSS14}, we do not remove features element-wise but always remove a full vector to simulate a spatial location or full layer being missing from the input. Hence, we also do not require to perform any rescaling of the features but rather mask the complete vectors from subsequent operations.

\subsection{Open-vocabulary saliency with DeViL}
 \emph{DeViL} can be used to obtain the probability of a given word or a phrase conditioned on vision features at layer $l$ and location $i,j$ by evaluating $p_{\text{LM}}(\boldsymbol{t}_{\text{query}}|f(g_l(x)_{i,j}))$. When passing overall features from a layer of interest, we obtain a likelihood of the query for every spatial location that we can visualize as a saliency map. By using a general purpose language model, there are no constraints on the query such that we can obtain saliency maps on concepts that lie outside the original training scope of the vision model. This is useful for seeing whether the vision features encode information about related concepts due to their correlation with training data or obtained as a side-effect.


\begin{table*}
    \centering
    \resizebox{0.80\textwidth}{!}{\begin{tabular}{lll|cccccc}
     Method & Vision Backbone & LM & B4 $\uparrow$ & M $\uparrow$ & RL $\uparrow$ & C $\uparrow$ & S $\uparrow$ & \#Params (M) $\downarrow$ 
     \\\hline
    \multirow{4}{*}{DeViL} & IN-ResNet50 & OPT & 5.851 & 9.572 & 23.92 & 65.73 & 15.23 & 88\\
    & CLIP-ResNet50 & OPT & 6.770 & 10.68 & 25.90 & 78.41 & 17.38 & 88\\
     
     & CLIP-ViT & OPT & \textbf{7.506} & \textbf{11.22} & \textbf{26.82} & \textbf{86.29} & \textbf{18.37} & 88
     \\ 
     & CLIP-ViT & GPT2 & 6.349 & 10.55 & 25.70 & 76.55 & 17.81 & 40
     \\
     ClipCap~\cite{mokady2021clipcap} & CLIP-ViT & GPT2 &
        - & - & 25.12 & 71.82 & 16.07 & 43
        \\  
     \hline 
     VLP~\cite{zhou2020unifiedvlp} & &  & 
        - &- & 24.35 & 77.57& 16.59 & 115
        \\ 
     LEMON~\cite{hu2022lemon} & & & \textbf{10.1} & \textbf{12.1} & - & \textbf{104.4} & \textbf{19.0} & 196.7 
 \end{tabular}}
    \caption{Image captioning results on CC3M~\cite{sharma2018cc3m} with different pre-trained vision backbones and language models, and a comparison with state-of-the-art captioning models, either fully~\cite{zhou2020unifiedvlp,hu2022lemon} or partially-finetuned~\cite{mokady2021clipcap}. We report standard captioning metrics, where higher is better. IN-ResNet50: ImageNet~\cite{deng2009imagenet} pre-trained ResNet50~\cite{he2016resnet}. CLIP-ResNet50/ViT: ResNet50/ViT versions of the CLIP~\cite{radford2021learning} vision encoder trained on the CLIP dataset.}
    \label{result_full}
\end{table*}

\section{Experiments}\label{exp_res}
\vspace{-2.5mm}
We evaluate \emph{DeViL} on both its natural language and saliency generation capabilities. To train \emph{DeViL} we use the CC3M~\cite{sharma2018cc3m}  captioning dataset. While our goal is to generalize to more fine-grained descriptions of vision features, \emph{DeViL} can still be used as a captioning model by itself. Therefore, we start by evaluating image captioning on CC3M, before discussing explanations of vision features obtained through \emph{DeViL}. 
For fine-grained analysis, we report both qualitative as well as quantitative results on fine-grained neuron descriptions by training and evaluating on the MILANNOTATIONS dataset~\cite{hernandez2022milan}. Lastly, we evaluate both NLEs and saliency obtained through \emph{DeViL} across different layers to show its generalization capabilities. As we focus on explaining the vision backbone instead of producing captions, all images used for qualitative analysis come from sources outside CC3M, such as ImageNet~\cite{deng2009imagenet}, Places365~\cite{zhou2017places}, and COCO~\cite{lin2014coco}. Details about the \emph{DeViL} architecture and training details can be found in the supplementary.

\subsection{Evaluating feature descriptions through image captioning}

\textbf{Dataset.} We use the official train-validation split of the CC3M~\cite{sharma2018cc3m}, consisting of 3M image-caption pairs collected from the web. We chose this dataset because it is sufficiently large to cover a large variety in both the vision and language modalities. Its successor CC12M~\cite{changpinyo2021cc12m} is less focused on high conceptual relevance, and more noisy in caption quality, making CC3M more suitable for tasks that require strong semantic alignment between text and image.

\textbf{Baselines.}Although our goal is to translate latent representations of pre-trained vision models into language, \emph{DeViL} can still be used to obtain full image descriptions. Hence, we evaluate \emph{DeViL} generated sentences with standard captioning metrics and compare against captioning methods~\cite{mokady2021clipcap,zhou2020unifiedvlp,hu2022lemon}. ClipCap~\cite{mokady2021clipcap} is a lightweight model that combines the CLIP vision model with a pre-trained language model. Similar to our approach, ClipCap only trains a translation network to keep the training cost low. Both UnifiedVLP~\cite{zhou2020unifiedvlp} and LEMON~\cite{hu2022lemon} are large-scale models pre-trained on several datasets not limited to captioning and subsequently finetuned. Thus, they surpass lightweight models such as ClipCap and \emph{DeViL}, but require a lot of resources to train.

\textbf{Results.}
\sloppy{We present our results on common language-based metrics: BLEU@4~\cite{papineni2002bleu}}, METEOR~\cite{denkowski2014meteor}, ROUGE-L~\cite{lin2004rouge}, CIDEr~\cite{vedantam2015cider}, and SPICE~\cite{anderson2016spice}. We consider ResNet50~\cite{he2016resnet} trained on ImageNet and CLIP~\cite{radford2021learning} in both its ResNet50 and ViT variants as our vision backbones.

We report image captioning results in Table~\ref{result_full}. Between vision backbones, we observe that CLIP-ViT performs better than its ResNet50 counterpart and ResNet50 trained on ImageNet, which is not surprising given CLIP's contrastive vision-language pre-training. Since the vision encoder of CLIP has in the past depicted strong zero-shot capabilities on a variety of tasks, we would expect it also to have a large coverage of visual concepts when we explain their features.

When using the CLIP-ViT backbone, we further ablate the relevance of the pre-trained language model. 
We make use of OPT-125M~\cite{zhang2022opt} and GPT2~\cite{radford2019gpt2} in our pipeline. With CLIP-ViT-B-32 as the vision backbone and OPT as the language model, we obtain our best scores surpassing ClipCap on all by a big margin, e.g. a CIDEr of 86.29 vs. 71.82. Even when using GPT2 and the same translation network architecture as ClipCap, we perform better across the board while using fewer parameters (40M vs. 43M). This is due to our model changes in using multiple layers of the vision backbone and the introduction of both feature-level and layer-level dropouts. Compared to large-scale captioning models that require more resources, we still perform better than UnifiedVLP on all metrics despite it requiring 1200h of training on a V100 GPU. In comparison, ClipCap reports 72h of training on GTX1080, while \emph{DeViL} requires 24h of training on an A100. LEMON~\cite{hu2022lemon} still surpasses our lightweight model as the state-of-the-art model on captioning. Required training resources for LEMON are not reported.

\begin{table}
    \centering
    \resizebox{\columnwidth}{!}{\begin{tabular}{ccc|ccccc}
     & Token & Feature & & & & & \\
     Layer & Dropout & Dropout & B4 $\uparrow$ & M $\uparrow$ & RL $\uparrow$ & C $\uparrow$ & S $\uparrow$\\\hline
     \multirow{3}{*}{single} & & &  0.3038 & 3.435 & 11.76 & 4.729 & 1.962
     \\
     & \checkmark & & 6.442 & 10.34 & 25.14 & 73.73 &  16.86 
     \\
     & \checkmark & \checkmark & 6.623 & 10.49 & 25.50 & 75.88 & 17.11 
     \\
     \hline
     \multirow{3}{*}{all} & & &  7.101 & 11.16 & 26.62 & 82.59 & 18.54 \\ 
      & \checkmark & & 7.211 & 11.09 & 26.70 & 83.32 & \textbf{18.48} \\ 
      & \checkmark & \checkmark & \textbf{7.506} & \textbf{11.22} & \textbf{26.82} & \textbf{86.29} & 18.37
     \\
 \end{tabular}}
    \caption{Ablating our dropout when evaluated using vision features from all layers or only the very last layer (single) of CLIP-ViT.} 
    \label{ablations}
\end{table}

\begin{figure*}[!ht]
  \includegraphics[width=\linewidth]{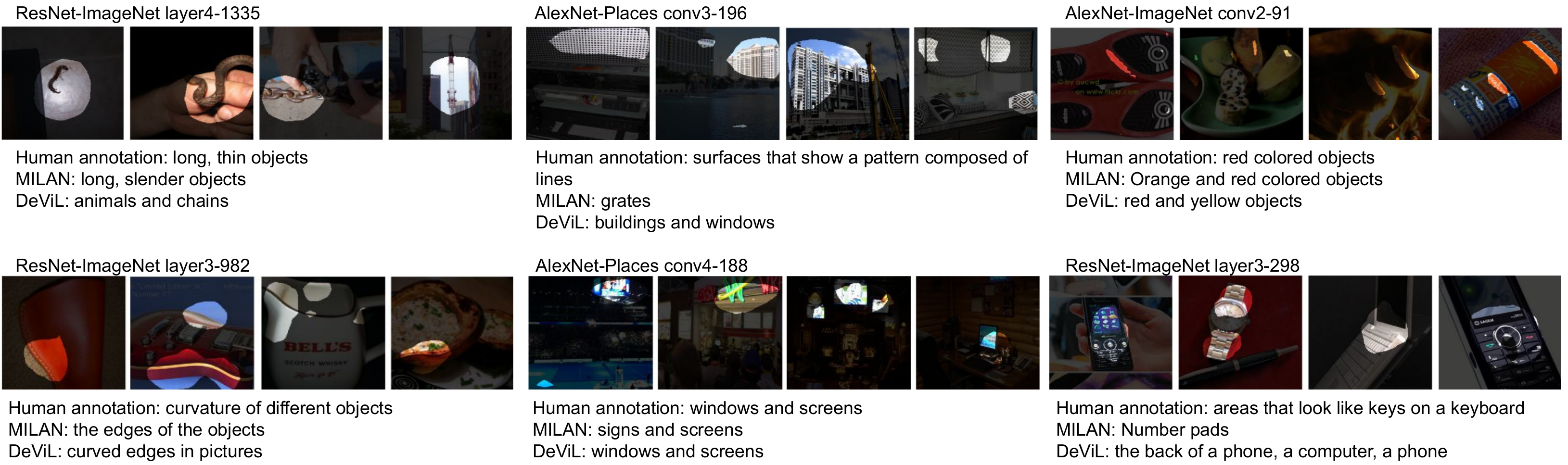}
  \caption{Qualitative results on MILANNOTATIONS~\cite{hernandez2022milan} and comparison with the MILAN~\cite{hernandez2022milan} model.}
  \label{fig:milan_res}  
\end{figure*}

\textbf{Ablations.} We ablate the proposed token and feature dropouts in Table~\ref{ablations} in terms of captioning metrics for the CLIP-ViT model. This table shows 3 different models trained with incremental combinations of both dropouts. While each \emph{DeViL} model is trained to explain multiple vision backbone layers, we evaluate them in two scenarios: using vision features from all layers of the vision backbone it was trained with (all) and when using only the last layer (single).
This ablation study confirms that dropout is essential for producing image-relevant captions for individual layers at inference time. Since the model without token dropout has never seen a single layer being passed to the translation network, it performs poorly in this out-of-distribution scenario making CIDEr drop from 75.88 to 4.729. Feature dropout is less essential, but further improves captioning scores and reduces overfitting. When we compare results on using all layers rather than just using one, we see an improvement in all scores, e.g. CIDEr increases from 75.88 to 86.29. This suggests that complementary information is encoded in the different layers and caption content indeed differs, making it reasonable to assume we can obtain layer-specific explanations even when training on caption data.

Overall, these results show that with our dropout methods, we can train on several layers at once and perform an evaluation on individual layers by themselves, while also avoiding the need to train one model for each layer we might want to explain later on.

\subsection{MILAN: explaining local visual features}
Since \emph{DeViL} was designed first and foremost for localized feature inspection in natural language, we strive to compare our method more directly on an explainability task, and especially on one such task at the local feature level. Thus, we also train \emph{DeViL} on MILANNOTATIONS~\cite{hernandez2022milan}, a dataset of over 50k human descriptions of sets of image regions obtained from the top neuron activating images of several known models like ResNet~\cite{he2016resnet} or AlexNet~\cite{krizhevsky2012imagenet}. For each base model, the authors collect descriptions for the top 15 activated images of each neuron in the network. Each image is masked by its corresponding activation mask, highlighting only the regions for which the corresponding neuron fired.

\begin{table}[t]
    \centering
    \resizebox{0.9\columnwidth}{!}{\begin{tabular}{l|cccccc}
     Method & B4 $\uparrow$ & M $\uparrow$ & RL $\uparrow$ & C $\uparrow$ & S $\uparrow$ & BS $\uparrow$ \\
     \hline
     MILAN~\cite{hernandez2022milan} & - & - & - & - & - &  0.362\\
     ClipCap~\cite{mokady2021clipcap} & 3.99  & 9.62  & 27.0 & 25.1 & 10.8 & 0.381 \\
     \emph{DeViL} & \textbf{6.28} & \textbf{11.3} & \textbf{30.6} & \textbf{33.7} & \textbf{13.3} & \textbf{0.382}\\
    \end{tabular}}
    \caption{Evaluating MILAN, ClipCap and \emph{DeViL} on MILANNOTATIONS.\vspace{-6mm}}
    \label{tab:milan}
\end{table}

We compare with ClipCap and the MILAN model~\cite{hernandez2022milan} trained on MILANNOTATIONS~\cite{hernandez2022milan}. Although our model was designed to work with layer-wise feature maps for a single image and not at the neuron level, we adapt \emph{DeViL} by pooling over the 15 masked images given for each neuron. We report the NLP metrics including BERTScore~\cite{zhangbertscore} results in Table~\ref{tab:milan}. The results are averaged over several generalization experiments proposed by \cite{hernandez2022milan}. We report a complete comparison in terms of BERTScore of all 13 experiments in the supplementary.

The average BERTScore across scenarios is 0.362, 0.381 and 0.382 for MILAN, ClipCap and \emph{DeViL}, respectively. The margins between all models are small as BERTScore is very sensitive to small differences that can still indicate a reliable ranking~\cite{zeiler2014visualizing}. Considering all other language metrics, the difference between ClipCap and \emph{DeViL} is more pronounced and \emph{DeViL} outperforms ClipCap consistently.
A qualitative comparison with MILAN~\cite{hernandez2022milan} can be seen in Figure~\ref{fig:milan_res}. We refer the reader to examples like ``animals and chains'' and ``building and windows''. Both quantitative and qualitative results validate \emph{DeViL's} generalization ability and its primary intended goal: to faithfully decode localized vision features into natural language.

\subsection{Diverse layer-wise explanations of vision features}
Deep neural networks learn to extract meaningful patterns from input data by progressively building up a hierarchy of features. The lower layers tend to detect simple patterns like edges and curves, while the higher layers learn to recognize more complex and abstract concepts~\cite{szegedy2015going,zeiler2014visualizing,agrawal2015learning,he2016resnet}. We verify a similar trend in the descriptions generated by \emph{DeViL}. We generate descriptions for each spatial location of the feature map at layer $l$ and produce saliency maps to measure the spatial support of the main subject in the sentence generated at a single location. Our model assigns a probability score to a textual query being generated based on its relevance to the visual features. This score can then be visualized as a heatmap, with higher scores corresponding to areas of the image where the vision model encoded the textual concept.

\begin{figure*}[t]
  \centering
  \includegraphics[width=.8\linewidth]{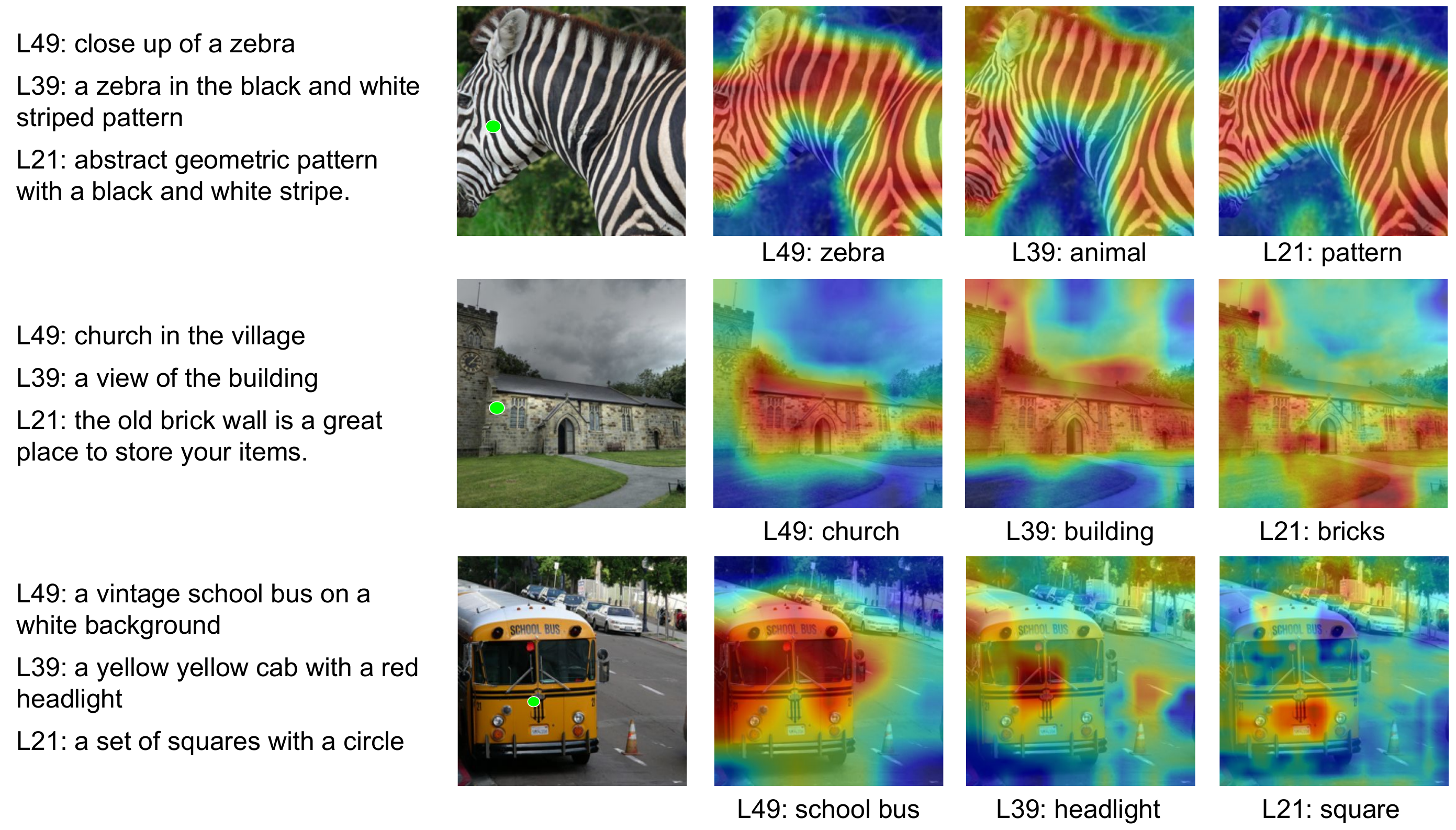}
  \caption{Captions generated for each image by the CLIP-ResNet50 model, for layer 49(L49), layer 39(L39) and layer 21(L21). To the right of original image, we present saliency maps for relevant words taken from the captions at the location marked with a green dot. With each layer the learned concepts go from higher level (more specific) to lower level (more general).}
  \label{fig:captions_cliprn50}
\end{figure*}

\textbf{Qualitative.} Figure~\ref{fig:captions_cliprn50} corresponds to the output generated by the CLIP-ResNet50 model for layers 21, 39, and 49. The green dot in the image shows the location for which the description has been generated. The generated descriptions showcase the aforementioned hierarchy, with lower layers mentioning lower level features like colors (e.g. ``red headlight'') and shapes (e.g. ``squares with a circle'') and higher layers mentioning objects (e.g. progression from ``building'' to ``church''). The saliency maps also validate the spatial location of these words, e.g. higher saliency scores for ``zebra'' at L49, ``animal'' at L39, and ``pattern'' at L21.

\begin{figure}[t]
    \centering
  \includegraphics[width=1.0\columnwidth]{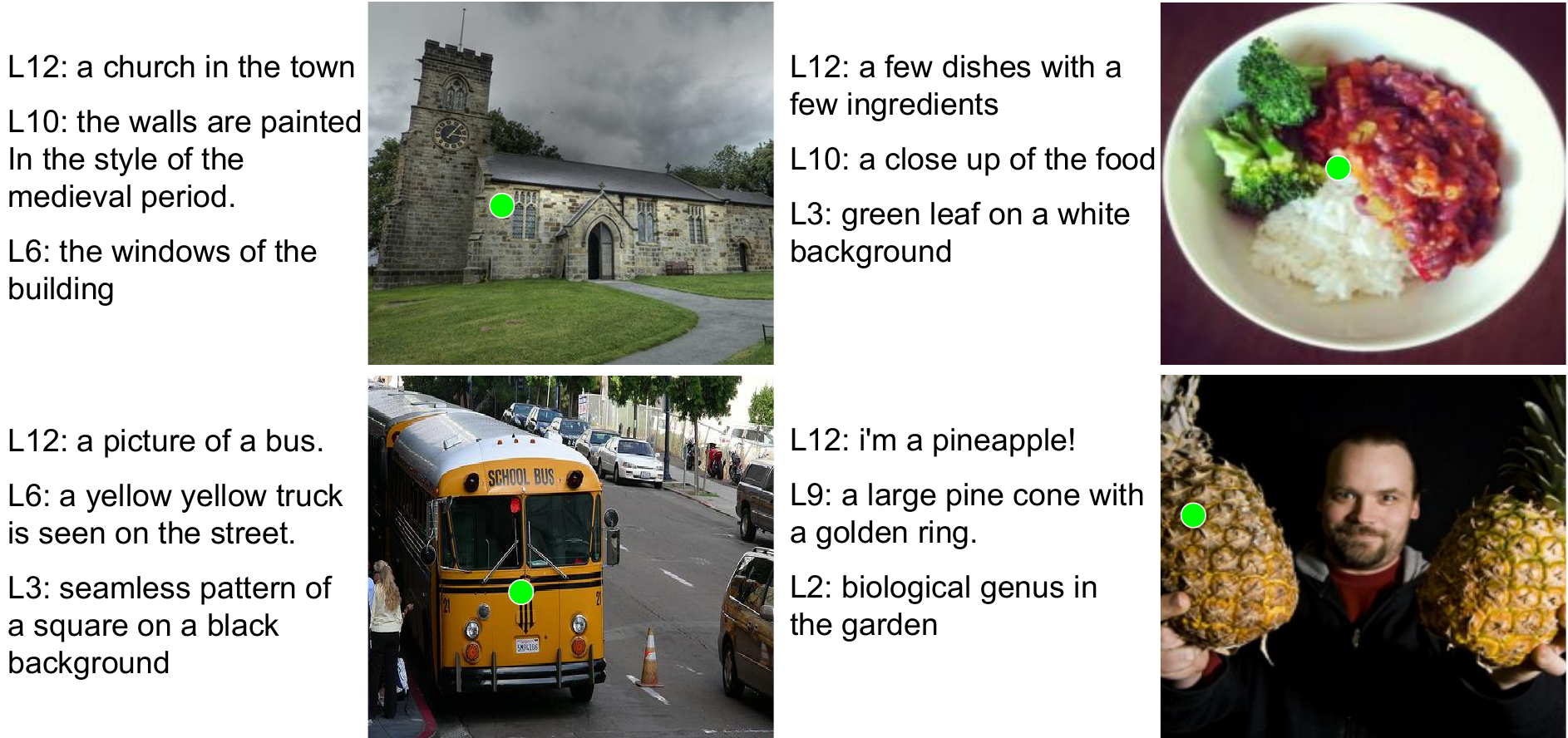}
  \caption{Descriptions generated for CLIP-ViT at different layers for the locations indicated by the green visual marker.}
  \label{fig:captions_clipvit}
\end{figure}

We also show natural language explanations generated by different layers of the Clip-ViT model. A key feature about ViTs is that they tend to generate similar feature maps in between blocks, which can be attributed to the way the self-attention mechanism hworks~\cite{dosovitskiy2020image,wang2021pyramid}. For this reason,  we evaluate the last 12 layers of transformer-based models in contrast to 3 layers in convolution-based models. As shown in the school bus image of Figure~\ref{fig:captions_clipvit}, CLIP-ViT generates the word ``bus'' in its last layer (L12), and moves to lower level concepts including the identification of the color ``yellow'' for L6, and lastly the shape ``square'' at L3. We try to capture a wide variety of objects in our qualitative analysis to show the generalization capability of the model.

\begin{table}[t]
    \centering
    \resizebox{0.9\columnwidth}{!}{\begin{tabular}{lccccc}
    \hline
        \multirow{2}{*}{Model} & \multirow{2}{*}{Layer} & Adj & Verbs & \#Uniq. &  CIDEr  \\ 
        & & (\%) & (\%) & Words & $\uparrow$ \\
        \hline
        CLIP-ViT & 9 & 6.37 & 6.30 & 4790 & 61.6  \\
        CLIP-ViT & 6 & 7.60 & 5.78 & 4115 & 44.1 \\ 
        CLIP-ViT & 3 & 8.04 & 5.24 & 2554 & 20.2 \\
        \hline
        CLIP-ResNet50 & 49 & 6.30 & 6.66 & 5228 & 71.2  \\ 
        CLIP-ResNet50 & 39 & 6.69 & 6.43 & 4395 & 54.1  \\ 
        CLIP-ResNet50 & 21 & 6.89 & 5.90 & 3374 & 34.6 \\ 
      \end{tabular}}
    \caption{Per-layer statistics of adjectives, verbs, \# of unique words.}\label{tab:caption_stats}
    \vspace{-5mm}
\end{table}

\begin{figure}[t]
    \centering
    \includegraphics[width=.8\linewidth]{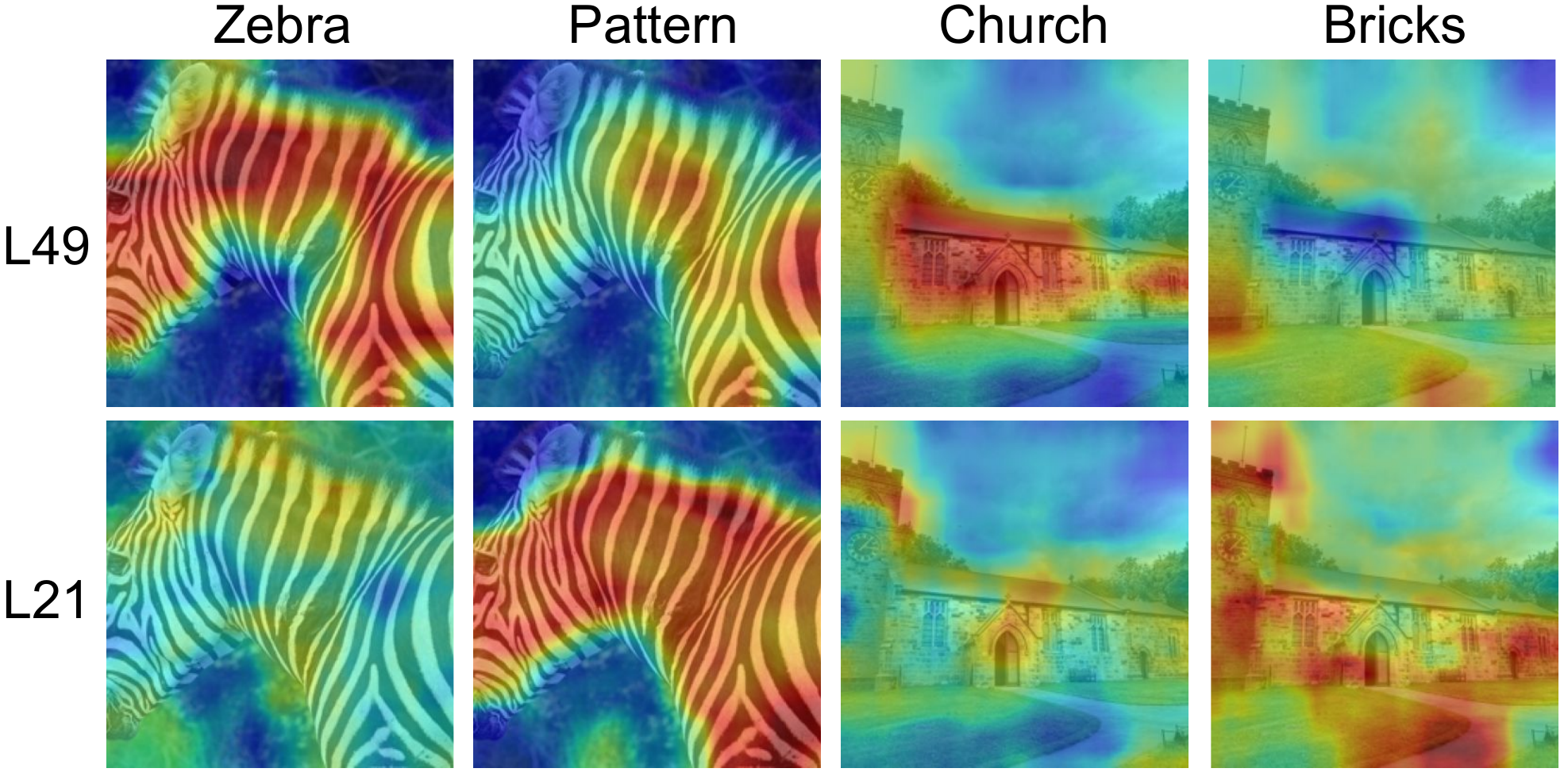}
    \caption{Semantics are encoded in different layers, CLIP-ResNet50 model}
    \label{fig:reb_sal_clip_rn50}
\end{figure}

\textbf{From simple to rich language descriptions.} To quantify how well \emph{DeViL} captures the differences of what specific vision layers encode, we analyze the generated text descriptions across layers. Specifically, we obtain a single text description per layer by averaging the vision features for each layer and decode them individually with \emph{DeViL}.
In Table~\ref{tab:caption_stats}, we take a look at the properties of the language generated by \emph{DeViL} on CC3M. As measured by the CIDEr score, we observe that the text similarity to human captions increases with later layers as more semantically meaningful embeddings are produced by the vision model.
We also see that the language shifts from using many adjectives in earlier layers to more verbs and a more comprehensive vocabulary in higher layers. This indicates a shift from describing simple concepts with a limited set of attributes to richer descriptions that also describe relations and actions between objects in the image. In Figure~\ref{fig:reb_sal_clip_rn50}, we show the saliency for the same word across different layers of CLIP-ResNet50. The top layer encodes ``Zebra'', but not ``Pattern'', and conversely for a lower layer. These insights can help us better understand intermediate features of a pre-trained model and assist in choosing the right features for a downstream task.

\begin{figure}[t]
  \centering
  \includegraphics[width=0.8\columnwidth]{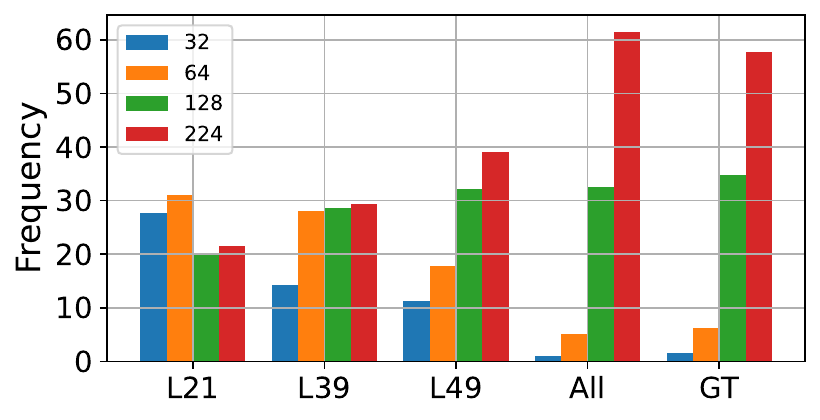}
  \caption{Frequency of most similar patch size for a given \emph{DeViL} description (L21, L39, L49, All) and human caption (GT) using CLIP-ResNet50.}
    \label{fig:clip-barplot}
    \vspace{-4mm}
\end{figure}

\textbf{Quantifying spatial and layer coherence.} We want to further analyze how well the spatially localized descriptions generated by \emph{DeViL} capture the content of the underlying image patches. Intuitively, earlier layers encode local information of a smaller underlying patch, while later layers can encode more global information. To test this, we create center crops of sizes 32x32, 64x64, and 128x128 along the full image size of 224x224 and compare their similarity to \emph{DeViL} descriptions of the vision feature closest to the center in each layer's feature map. We use CLIP-ResNet50 to embed the cropped images as well as \emph{DeViL} descriptions and compare their similarities. Figure~\ref{fig:clip-barplot} plots which crop size is most similar with the generated text. We observe that smaller patches have much higher similarity with earlier layer descriptions, and conversely bigger patches with later layers. Descriptions of L21 best match 64x64 patches and this distribution shift progressively to L49 fitting best to the full image content. These results validate that \emph{DeViL} exposes what the vision model encodes both in lower layers describing the content of local patches and global concepts in higher layers.

\begin{figure}[t]
  \centering
  \includegraphics[width=0.85\linewidth]{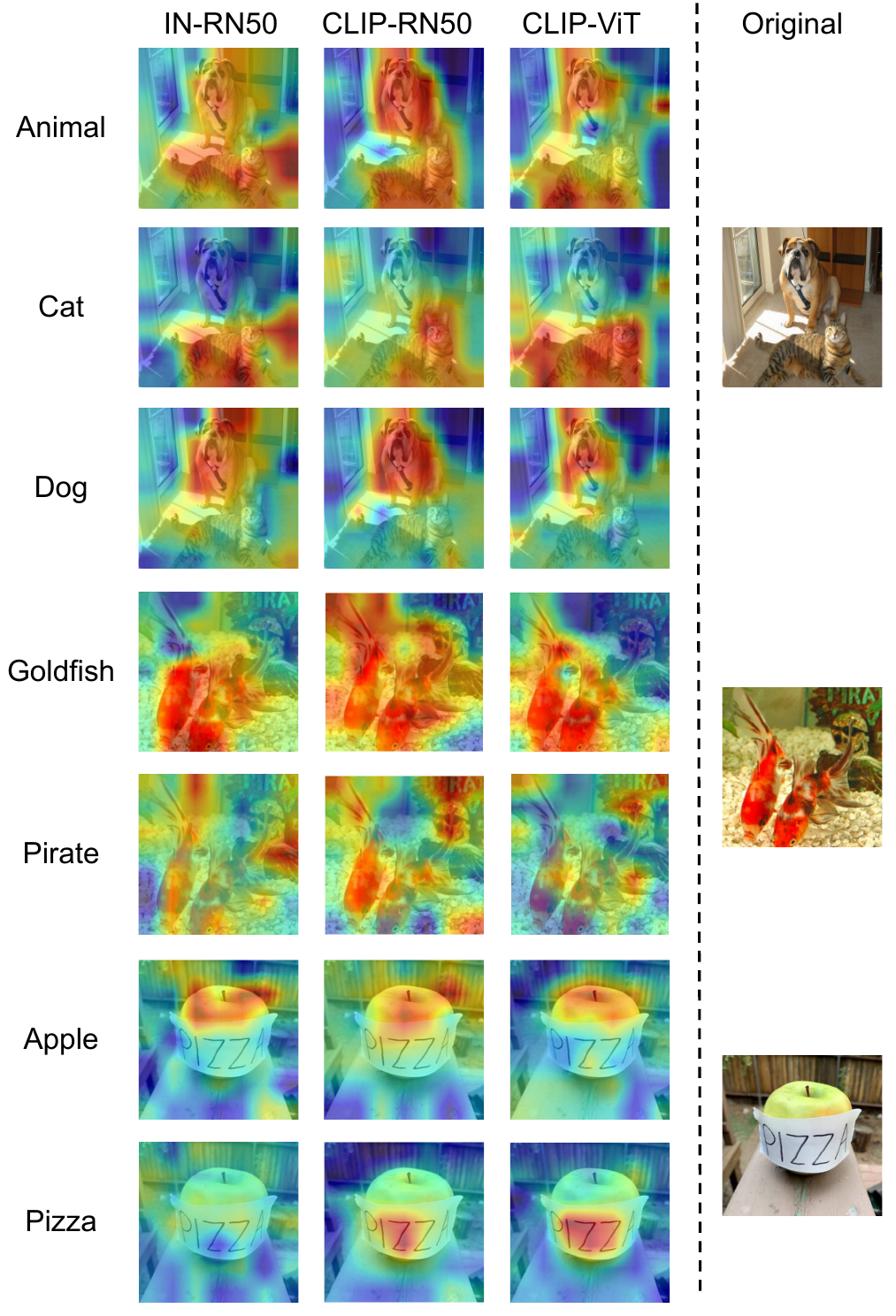}
  \caption{Open-vocabulary saliency maps for different backbones with the final layer. For CLIP-ViT we use the penultimate layer.}
  \label{fig:saliency}
\end{figure}

\subsection{Inspecting different vision models through saliency}
Since \emph{DeViL} incorporates an LLM, it has the ability to produce saliency maps for any word, not being limited by the closed-set of classes of a given dataset. Figure~\ref{fig:saliency} presents a comparison of different vision backbones in terms of saliency maps for different words. The different models can distinguish ``cat'' and ``dog'', but also identify both as ``animal''. Interestingly, since CLIP-based models are trained with not only images but also text, they can identify the written word ``pizza'' on the bottom-most image, while an ImageNet trained ResNet50 cannot. Similarly, the word ``pirate'', written on a stone, is also identified by CLIP-based models for the aquarium image. These failure cases of the CLIP model have been discovered independently and the explanations of our \emph{DeViL} model expose these as well, highlighting its faithfulness to the vision model. Obtaining saliency maps for open-vocabulary words that did not appear in a model's training data such as the supercategory ``animal'' or the class ``pirate'' is a novel contribution of our method. It also allows a more direct comparison with models of different pretraining tasks such as CLIP, which is not possible with methods that rely on the model's output to be the same for direct comparisons. More qualitative results can be found in the supplementary material.

We also compare our saliency maps quantitatively with Grad-CAM~\cite{selvaraju2017grad}, which proposes interpreting a neural network via the gradients for a target class flowing through a given layer. Results are presented in Table~\ref{saliency_metrics} for 3 layers of the ResNet50 model (L49, L39, and L11). We report commonly used deletion (Del.) and insertion (Ins.) metrics~\cite{PetsiukDS18}. The Del.(Ins.) metric measures the drop(rise) in the probability of a class as relevant pixels given by the saliency map are gradually removed(inserted) from the image. Thus, lower(higher) is better for Del.(Ins.).
Although we are unable to outperform Grad-CAM, the latter uses gradients, which provide more information than just activations alone. Furthermore, Grad-CAM is specialized in the network's output classes, while \emph{DeViL} can perform open-vocabulary saliency. An interesting future research direction is to improve and align open-vocabulary saliency with existing approaches.

\begin{table}[t]
   \centering
   \resizebox{0.85\columnwidth}{!}{
    \begin{tabular}{ll|cc}
    Layer & Metric & Grad-CAM~\cite{selvaraju2017grad} & DeViL (CC3M)\\
    \hline
    \multirow{2}{*}{L49} 
    & Del. & 0.342 & 0.466\\
    & Ins. & 0.763 & 0.697\\
    \hline
    \multirow{2}{*}{L39}
    & Del. & 0.465 & 0.494\\
    & Ins. & 0.677 & 0.661\\
    \hline
    \multirow{2}{*}{L21}
    & Del. & 0.405 & 0.484\\
    & Ins. & 0.593 & 0.635\\ 
    \end{tabular}
    }
    \caption{Comparison between the ResNet50 model and Grad-CAM~\cite{selvaraju2017grad} for saliency maps generated for different layers (L49, L39, and L21), in terms of deletion ($\downarrow$) and insertion ($\uparrow$) metrics.}
    \label{saliency_metrics}
\end{table}

\section{Conclusion}
The ability to interpret deep learning models' predictions is crucial for their effective deployment in real-world applications. The visualizations of intermediate feature maps have been shown to be a promising approach for improving model interpretability, but understanding these feature maps often requires further analysis. Our proposed \emph{DeViL} approach for explaining feature maps through natural language is unique as it generates textual descriptions for individual features, and can also produce open-vocabulary saliency attributions. We evaluate the efficacy of our model's language generations on CC3M and MILANNOTATIONS, outperforming competing models, and show extensive qualitative results validating that our open-vocabulary saliency exposes which concepts are understood by each of the model's layers.

\subsection*{Acknowledgments}
This work was supported by the ERC (853489-DEXIM), DFG (2064/1 – project number 390727645), BMBF (Tübingen AI Center, FKZ: 01IS18039A), FCT (under PhD grant 2020.07034.BD), Carl Zeiss Stiftung and Else Kröner-Fresenius-Stiftung (EKFS). The authors thank the International Max Planck Research School for Intelligent Systems (IMPRS-IS) for supporting  Meghal Dani.

{\small
\bibliographystyle{ieee_fullname}
\bibliography{egbib}
}

\newpage
\appendix
\title{\emph{DeViL:} Decoding Vision features into Language\\-\\Supplementary Material}
\author{}
\maketitle

\section{Implementation details}
The translation network $f$ uses the standard Transformer architecture~\cite{VaswaniSPUJGKP17} with $12$ layers. Across all experiments, we use $10$ learnable tokens for the prefix $\boldsymbol{p}$, as per the ablation study done by Mokady~\etal~\cite{mokady2021clipcap}, since we also did not observe an improvement beyond this number. For models trained with token and feature dropout, the dropout probability is $0.5$. Since the input to $f$ differs significantly between inference and training, a high dropout rate is beneficial. We further ensure that during training we never dropout all spatial locations or all layers for a given input, and re-sample the dropout mask in such cases.

For all experiments we use the AdamW~\cite{loshchilovdecoupled} optimizer, with learning rate $=1e-4$, weight decay $=1e-6$, and a linearly decaying scheduler with a minimum learning rate of $1e-6$ and $5k$ warmup steps. The batch size is $64$ and we limit generation to $50$ tokens ($20$ tokens for MILANNOTATIONS experiments). Since the OPT-125M~\cite{zhang2022opt} language model works best on the captioning task, we use it for all other experiments and qualitative results. In our ablation with GPT2~\cite{radford2019gpt2}, where we strive to directly compare against ClipCap~\cite{mokady2021clipcap}, we make the following changes to our translation network: lower the number of layers and attention heads from 12 to 8, and use an intermediate size of 1532 instead of the original 3072. This way we match the network hyperparameters of ClipCap. Our network ends up having fewer parameters (40M vs. 43M), since ClipCap uses a large linear layer to expand the CLIP embedding to 10 tokens while we only use 3 tokens, one per explained layer. Thus, we also achieve 35\%-40\% lower run time than ClipCap.

For the ResNet50 vision backbones, we explain the activations of the last layer of each residual block. The reported explanations, e.g. for layer L49, L39, L21, indicate the layer number counting all convolutional and linear layers, but correspond to a residual block each. For the ViT model, we explain each transformer layer after its feedforward projection, i.e. CLIP-ViT consists of a total of 12 transformer layers. For the MILANNOTATIONS experiments, we use CLIP-ResNet50 as well as the same ResNet101 vision backbone and explanation layers as in~\cite{hernandez2022milan}, which also use the activations after each residual block.

\begin{figure}[!ht]
\begin{center}
    \includegraphics[width=0.9\linewidth]{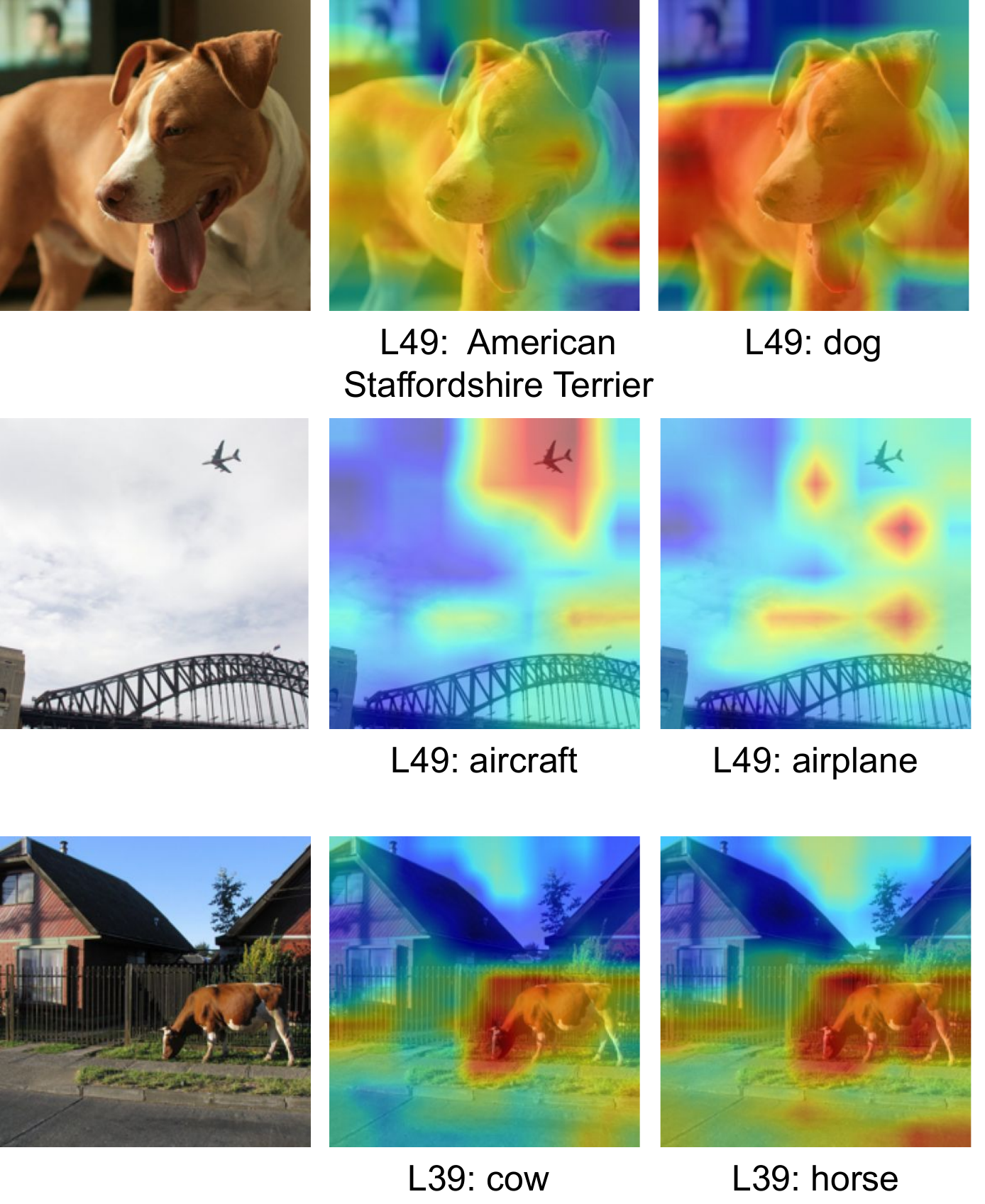}
\end{center}
   \caption{Failure cases of \emph{DeViL} for CLIP-RN50. Examples include the inability to identify specific dog breeds and specific words such as "airplane". We hypothesize that these limitations might be inherited from the limitations of the training data vocabulary of CC3M.}
\label{fig:limitations}
\vspace{-5mm}
\end{figure}

\begin{figure*}[!ht]
\begin{center}
    \includegraphics[width=0.9\linewidth]{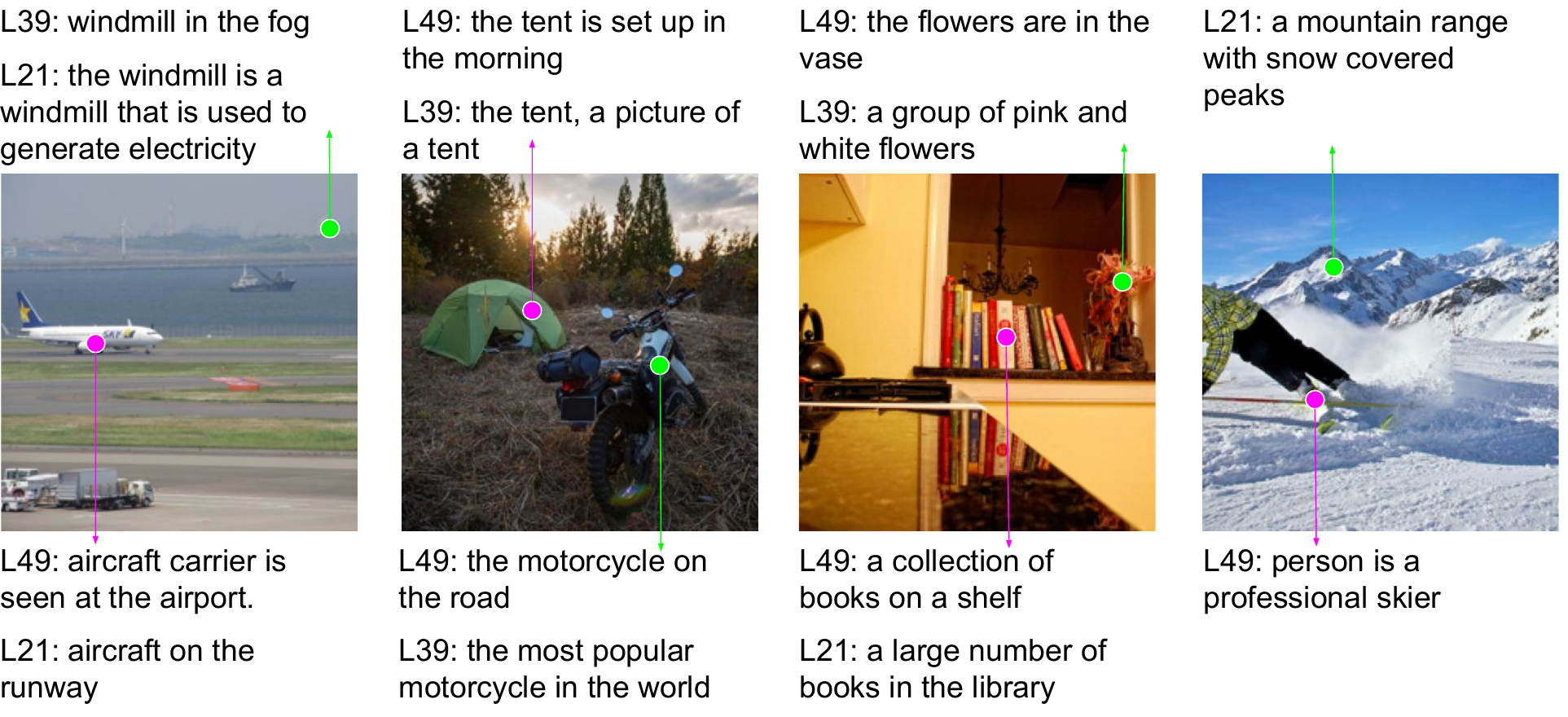}
\end{center}
   \caption{Additional examples of generated localized natural language explanations for the CLIP-ResNet50 vision backbone on images from the COCO dataset. For each image we show two locations being explained (indicated by the pink and green markers). For each location, we present explanations from different layers of model being explained (L49, L39, etc), where a higher number corresponds to a layer closer to the output of the network.}
\label{fig:cap_cliprn50_coco}
\end{figure*}

\section{Limitations and failure cases}
\begin{figure*}[t]
\begin{center}
    \includegraphics[width=0.65\linewidth]{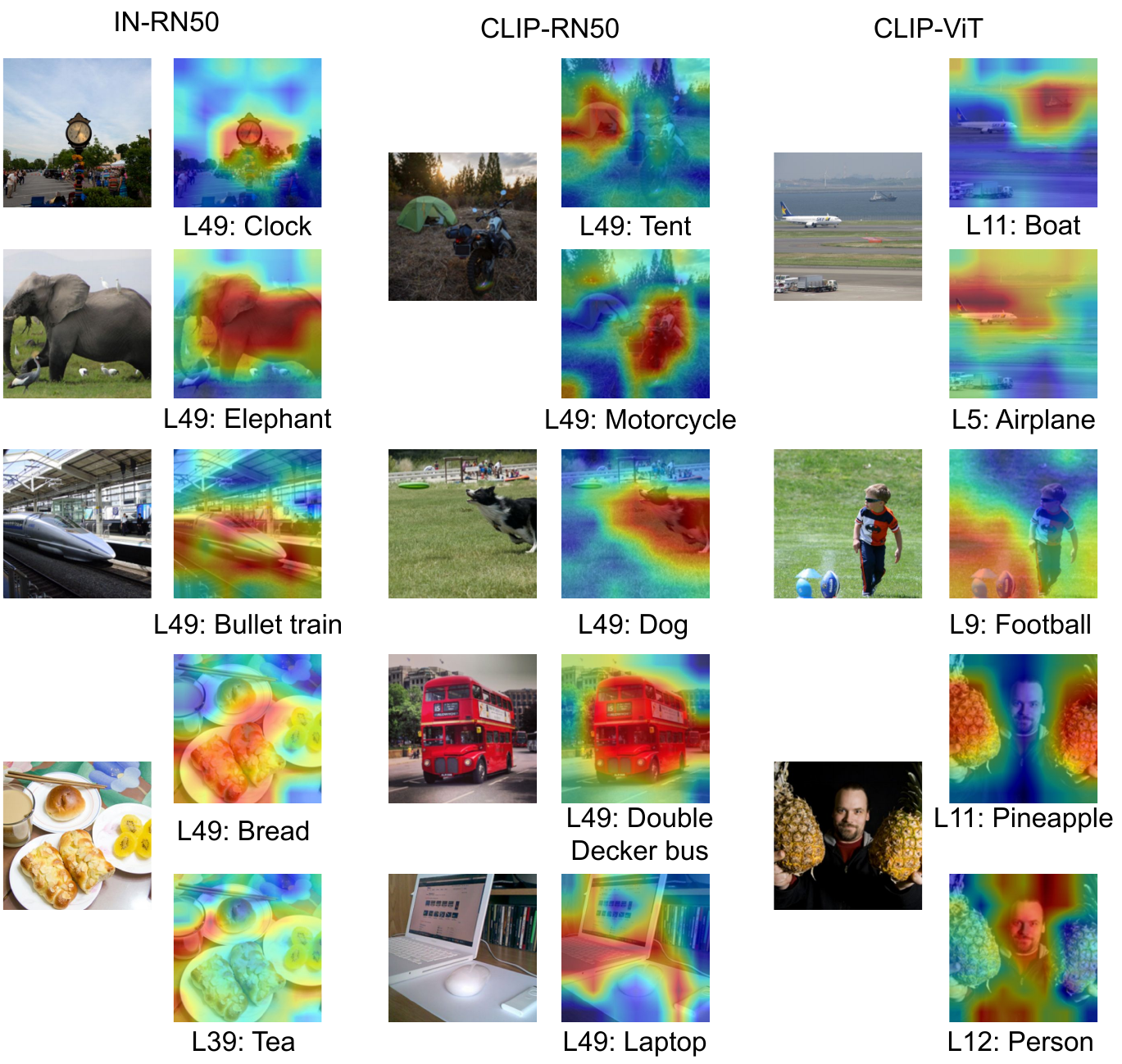}
\end{center}
   \caption{Additional open-vocabulary saliency examples for different vision backbones (IN-RN50: ResNet50 pre-trained on ImageNet, CLIP-RN50, and CLIP-ViT). For each example, we present the original image alongside saliency maps generated for different words and different layers of the models being explained (L49, L11, etc). A higher layer number corresponds to a layer closer to the output of the network.}
\label{fig:saliency_all}
\end{figure*}

We use the CC3M~\cite{sharma2018cc3m} dataset for training our \emph{DeViL} model. Even though the data covers a large variety of concepts, some words or phrases can be missing from the text data. Our model struggles to recognize more specific class descriptions, such as dog breeds like ``American Staffordshire Terrier'', as they appear less frequently in the data. However, it can easily recognize more general classes, like ``dog'', as shown in Figure~\ref{fig:limitations}. Since CLIP-RN50 has a 44.45\% confidence of the image showing an ``American Staffordshire Terrier'' when evaluating CLIP's zero-shot classification over the imagenet classes, we can conclude that this is indeed a failure case of our explanation model and not of the vision backbone.
We also observed that the saliency of the model's predictions can differ for synonyms. For instance, when we compared ``aircraft'' and ``airplane'' in the second image, the saliency for ``aircraft'' was highly localized on the object, while it was not the case for ``airplane''. However, our model correctly identifies ``airplane'' in other images, e.g. in Figure~\ref{fig:saliency_all}.
Furthermore, when the vision backbone confuses similar features the same can be observed in our model's explanations. In the last row of Figure~\ref{fig:limitations}, the region of the animal activates highly for both ``horse'' and ``cow''. When we directly contrast ``cow'' and ``horse'' in zero-shot classification, CLIP-RN50 recognizes ``horse'' with a confidence of 66.65\% and cow with 33.35\% for the given input image. Exploring these limitations, we think that training our model on more diverse and larger data can be a potential solution to address this issue, thereby improving the reliability of the model.

\begin{figure*}[!ht]
\begin{center}
    \includegraphics[width=0.9\linewidth]{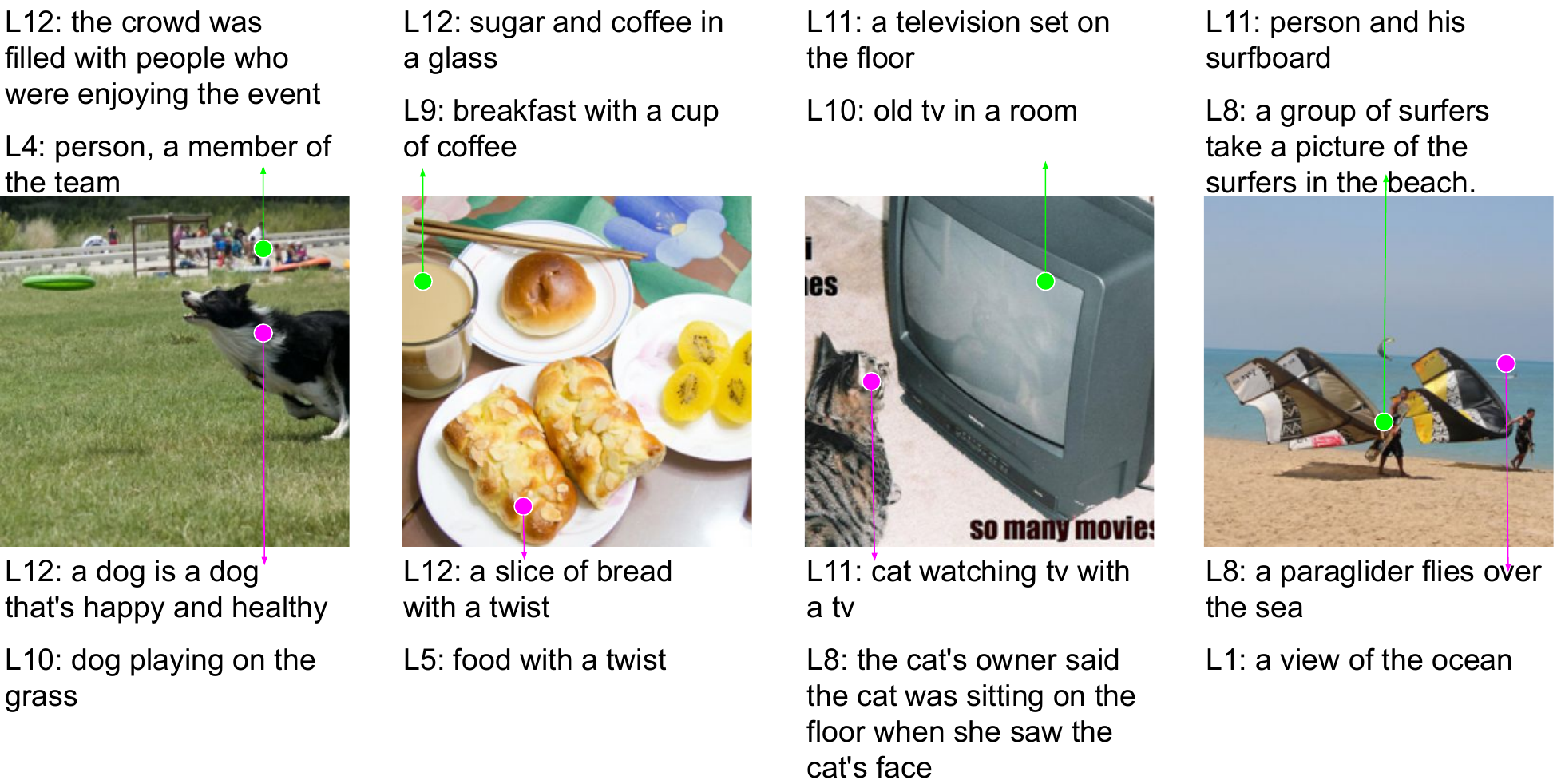}
\end{center}
   \caption{Additional examples of generated localized natural language explanations for the CLIP-ViT vision backbone on images from the COCO dataset. For each image we show two locations being explained (indicated by the pink and green markers). For each location, we present explanations from different layers of model being explained (L12, L5, etc), where a higher number corresponds to a layer closer to the output of the network.}
\label{fig:cap_clipvit_coco}
\end{figure*}

\begin{figure*}[!ht]
\begin{center}
    \includegraphics[width=0.9\linewidth]{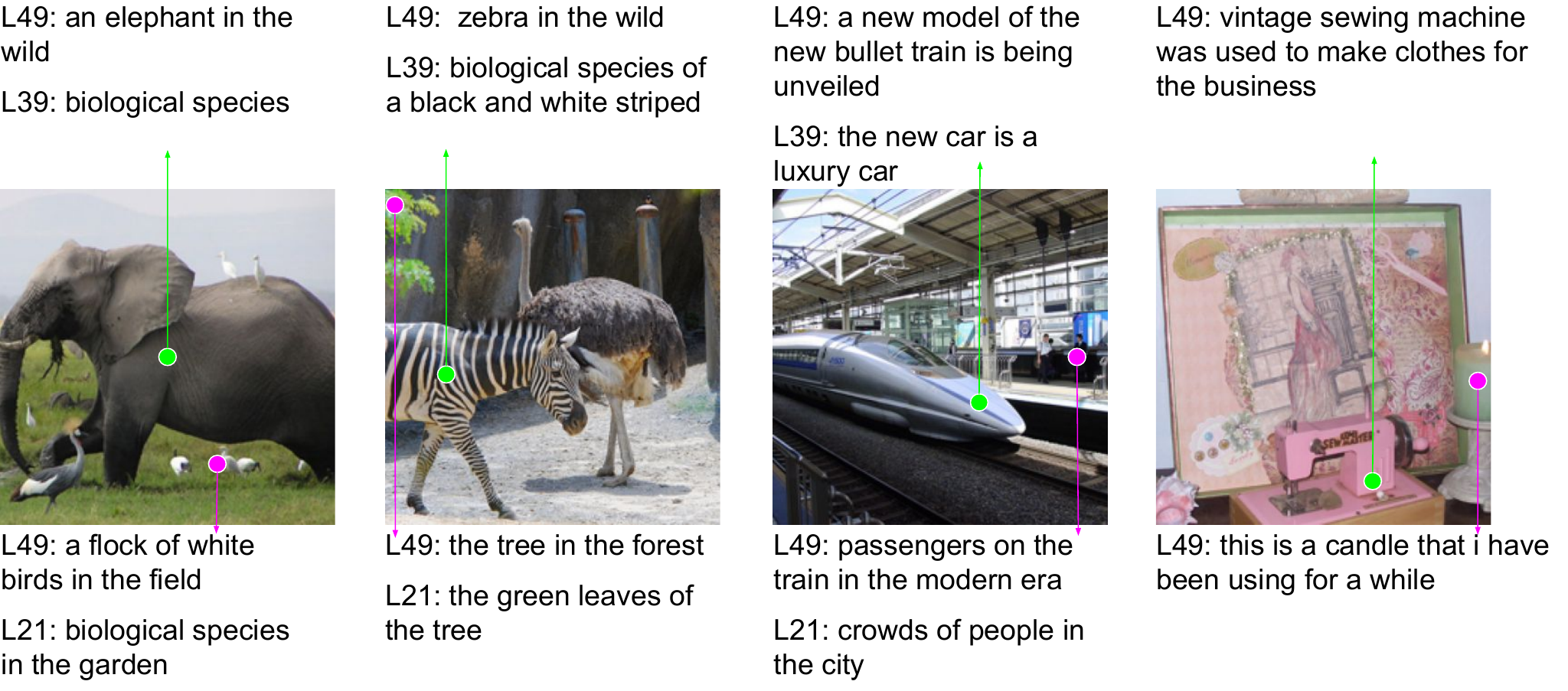}
\end{center}
   \caption{Additional examples of generated localized natural language explanations for the CLIP-ResNet50 vision backbone on images from the ImageNet dataset. For each image we show two locations being explained (indicated by the pink and green markers). For each location, we present explanations from different layers of model being explained (L49, L21, etc), where a higher number corresponds to a layer closer to the output of the network.}
\label{fig:cap_cliprn50_in}
\end{figure*}

\section{Feature descriptions at different locations}
We present additional qualitative results to showcase the efficacy of our proposed model \emph{DeViL}. It is not only capable of explaining the different layers of the vision model being inspected but maintains the spatial arrangement of features providing meaningful descriptions at different locations of the feature map. We test our pipeline on different vision backbones and datasets. In this section, we present an analysis for CLIP-ResNet50 and CLIP-ViT-B-32. We show different layer outputs at two different locations using green and pink visual markers on the image. In Figure~\ref{fig:cap_cliprn50_coco}, our model is able to detect a ``windmill'' (marked in green) and an ``aircraft'' (marked in pink). It is also able to generate meaningful captions for the lower layers of the vision backbone. For example, for the third image, the presence of flowers is noted on the final layer (L49), and in the previous layer (L39) the pink and white colors of the flowers are recognized. Similarly, we also present the output for a variety of objects present in the COCO~\cite{lin2014coco} using CLIP-ViT and ImageNet~\cite{deng2009imagenet} using CLIP-RN50 in Figures~\ref{fig:cap_clipvit_coco} and ~\ref{fig:cap_cliprn50_in} respectively.

\begin{figure*}[ht]
\begin{center}
    \includegraphics[width=1.0\linewidth]{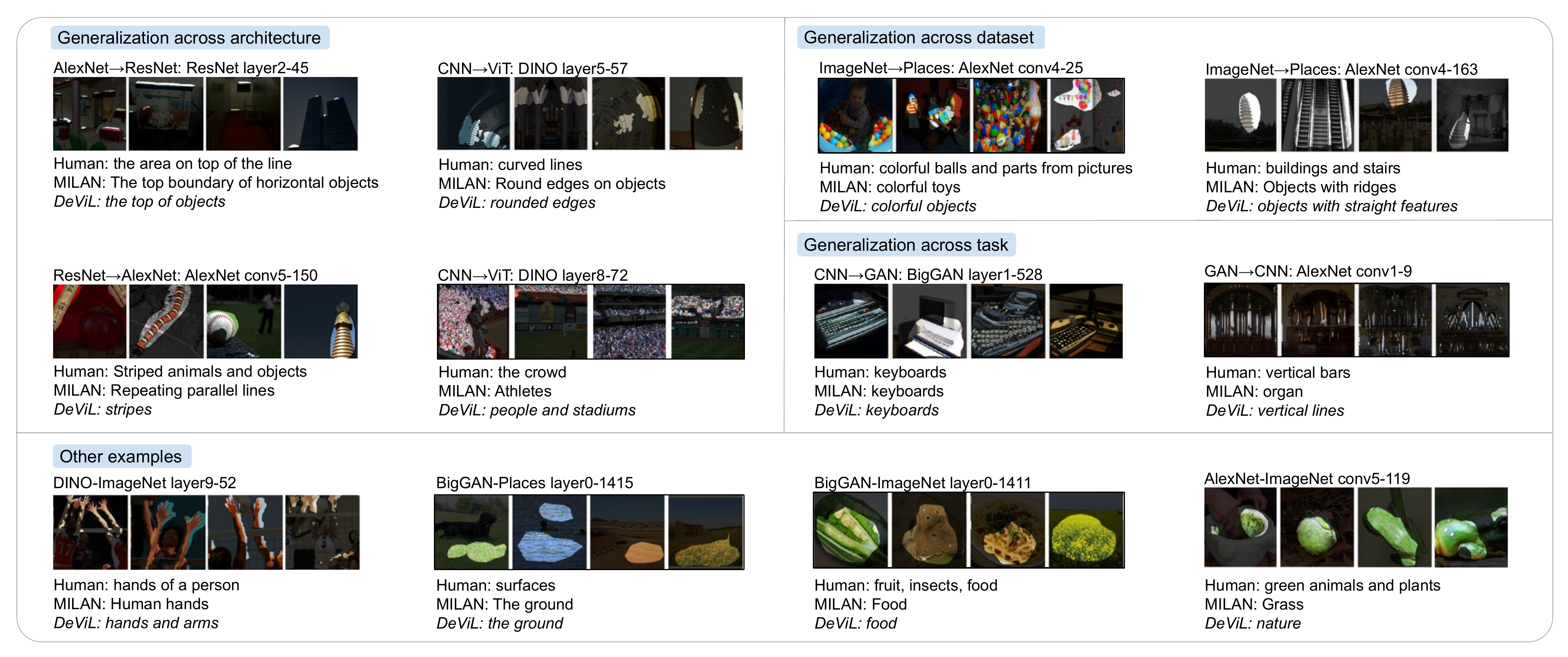}
\end{center}
   \caption{Additional qualitative examples on the MILANNOTATIONS dataset and comparison with the MILAN~\cite{hernandez2022milan} model. For each image in the generalization blocks, we specify the generalization experiment (e.g. AlexNet\textrightarrow ResNet) before the model, layer and neuron being explained.}
\label{fig:milan}
\end{figure*}

\begin{table*}[t]
   \centering
   \resizebox{0.8\linewidth}{!}{
    \begin{tabular}{lll|cccc}
    Generalization & Train & Test & MILAN~\cite{hernandez2022milan} & ClipCap~\cite{mokady2021clipcap} & \emph{DeViL} (RN101) & \emph{DeViL} (CLIP-RN50)\\
    \hline
    \multirow{6}{*}{Within network}
    & \multicolumn{2}{l|}{AlexNet-ImageNet} & 0.39 & \textbf{0.43} & \textbf{0.43} & \textbf{0.43}\\
    & \multicolumn{2}{l|}{AlexNet-Places} & \textbf{0.47} & \textbf{0.47} & 0.46 & 0.46\\
    & \multicolumn{2}{l|}{ResNet152-ImageNet} & 0.35 & \textbf{0.38} & \textbf{0.38} & \textbf{0.38} \\
    & \multicolumn{2}{l|}{ResNet152-Places} & 0.28 & 0.33 & \textbf{0.37} & 0.33\\
    & \multicolumn{2}{l|}{BigGAN-ImageNet} & 0.49 & \textbf{0.51} & 0.49 & \textbf{0.51}\\
    & \multicolumn{2}{l|}{BigGAN-Places} & 0.52 & 0.50 & 0.53 & \textbf{0.54}\\
    \hline
    \multirow{3}{*}{Across arch.}
    & AlexNet & ResNet152 & 0.28 & \textbf{0.33} & 0.30 & 0.30\\
    & Resnet152 & AlexNet & 0.35 & 0.35 & \textbf{0.36} & \textbf{0.36}\\
    & CNNs & ViT & 0.34 & \textbf{0.38} & 0.37 & 0.37\\
    \hline
    \multirow{2}{*}{Across datasets}
    & ImageNet & Places & 0.30 & 0.32 & \textbf{0.33} & \textbf{0.33}\\
    & Places & ImageNet & \textbf{0.33} & 0.31 & \textbf{0.33} & 0.32\\
    \hline
    \multirow{2}{*}{Across tasks}
    & Classifiers & BigGAN & 0.34 & \textbf{0.37} & 0.36 & 0.35\\
    & BigGAN & Classifiers & 0.27 & 0.27 & 0.28 & \textbf{0.29}\\
    \hline
    \multicolumn{3}{l|}{Mean} & 0.362 & 0.381 & \textbf{0.384} & 0.382
    \end{tabular}
    }
    \caption{Results on several generalization experiments with the MILANNOTATIONS~\cite{hernandez2022milan} dataset, in terms of BERTScore (higher is better).
    }
    \label{results_milan}
\end{table*}

\section{Saliency maps across vision backbones}
In this work, we not only work on Natural Language Explanations (NLEs) but also produce open-vocabulary saliency maps corresponding to overall features from a specific layer of interest. In Figure~\ref{fig:saliency_all}, we present a variety of words being recognized for different scenarios. We compare different vision backbones including ResNet50 pre-trained on ImageNet (IN-RN50), CLIP-ResNet50 (CLIP-RN50), and CLIP-ViT-B-32 (CLIP-ViT). We can observe that the CLIP-ViT model is able to detect objects such as ``pineapple'' and ``boat'' in the penultimate layer. While CLIP-RN50 and RN50 recognize most of the fine-grained objects in the last layer including ``tent'' and ``elephant''.

\section{Detailed comparison with MILAN}
Figure~\ref{fig:milan} presents additional qualitative results on the MILANNOTATIONS dataset and a comparison with the MILAN~\cite{hernandez2022milan} model. As seen by the different generalization experiments, \emph{DeViL} is able to generalize well and perform better than MILAN. Our model is able to correctly describe neurons using diverse language for models trained across new architectures, new datasets and training objectives. This is evident in the ``Generalization across architecture'' block in Figure~\ref{fig:milan}. It shows a peculiar case where \emph{DeViL} is able to identify ``people and stadiums'', which is a failure case reported by Hernandez et al~\cite{hernandez2022milan}. Our method also produces competitive results for the ``Other Examples'' block, where the caption produced for the leftmost case is ``hands and arms''. Unlike human annotations and MILAN, \emph{DeViL} not only mentions hands, but also arms, which are also visible in the masked images.

We also compare quantitatively with the state-of-the-art network dissection MILAN model~\cite{hernandez2022milan} and with ClipCap~\cite{mokady2021clipcap}. Although our model was designed to work with layer-wise feature maps for a single image and not at the neuron level, we adapt \emph{DeViL} by pooling over the 15 masked images given for each neuron. We report the BERTScore~\cite{zhangbertscore} results in Table~\ref{results_milan} (higher is better), on several of the generalization experiments proposed by the MILAN~\cite{hernandez2022milan} authors. For fair comparison, both ClipCap and DeViL use CLIP-RN50 as the vision backbone in Table~\ref{tab:milan}. In Table~\ref{results_milan}, we additionally report the results using ResNet101 (RN101) as vision backbone for DeViL as it was used by MILAN~\cite{hernandez2022milan}. We are able to outperform both MILAN~\cite{hernandez2022milan} and ClipCap~\cite{mokady2021clipcap} on mean BERTScore with either backbone.

\end{document}